\documentclass[10pt,twocolumn,letterpaper]{article}

\usepackage[pagenumbers]{cvpr}   
\renewcommand{\paragraph}[1]{\par\vspace{.5em}\noindent\textbf{#1.}\ }

\usepackage[utf8]{inputenc}
\usepackage[T1]{fontenc}
\usepackage{amsmath}
\usepackage{amssymb}
\usepackage{graphicx}
\usepackage{booktabs}

\definecolor{cvprblue}{rgb}{0.21,0.49,0.74}
\usepackage[breaklinks,colorlinks,allcolors=cvprblue]{hyperref}

\def\paperID{0000}
\def\confName{CVPR}
\def\confYear{2026}

\title{Automatic weld seam segmentation for industrial quality control:
a comparison of RGB and polarimetric imaging with CNN and
transformer architectures}

\author{Simone Garbin \quad Leonardo Venturoso\thanks{Corresponding author:
\texttt{leonardo.venturoso@fraunhofer.it}} \quad Marco Todescato\\
Fraunhofer Italia Research, Via A.\ Volta 13/A, 39100 Bolzano, Italy\\
{\tt\small\{simone.garbin, leonardo.venturoso, marco.todescato\}@fraunhofer.it}
}

\begin{document}
\maketitle

\begin{abstract}
Visual inspection of welded assemblies remains one of the least automated
stages in many industrial production processes, still depending largely on
the experience of human operators and thus subject to inter-operator
variability; the manufacturing of special-purpose machinery cabins, the
setting of this study, is one representative case. This work evaluates the
feasibility of automatic weld seam segmentation from RGB and polarimetric
imagery, comparing controlled laboratory acquisitions with images captured
under real, uncontrolled conditions. Convolutional neural network (CNN)
architectures (YOLOv8/YOLOv11 segmentation variants) and transformer-based
architectures (RF-DETR-Seg, Mask2Former) are benchmarked under a unified,
threshold-independent protocol, training each CNN with three random seeds to
separate genuine effects from seed noise. In controlled RGB conditions, CNN
models reach a mean mask mAP\textsubscript{50} (averaged over three training
seeds) of up to 0.87, but drop to 0.22--0.48 under uncontrolled acquisition,
showing that the acquisition setup is a first-order component of the
inspection system. Polarimetric imaging with alignment-preserving geometric
augmentation localizes previously unseen welds with a mean mask
mAP\textsubscript{50} up to 0.93: on par with, rather than ahead of, the best
controlled-RGB result, but reaching that accuracy on uncontrolled RGB without
requiring acquisition control. The clearest architectural finding concerns
robustness to viewpoint change. In-distribution, transformers and CNNs are
broadly comparable; but when the camera viewpoint shifts at test time
(zero-shot transfer to a close-range, robot-relevant viewpoint), the
transformer models, and RF-DETR in particular, retain high accuracy while
every CNN collapses. This gap holds across three training seeds and survives
a resolution-matched control, indicating it reflects the architecture rather
than training resolution. Within the CNN family, in contrast, model capacity
gives no reliable in-distribution advantage once seed variance is accounted
for. Small, fast CNNs are therefore a safe default when the deployment
viewpoint matches training, whereas transformer-based models deserve
consideration wherever that viewpoint may vary.
\end{abstract}


\section{Introduction}
\label{sec:intro}

Quality control of welded assemblies remains one of the least automated stages in the manufacturing process~\cite{imad2022intelligent}. Post-weld inspection is typically performed visually by specialized operators, a process that is time-consuming, inconsistent, and sensitive to environmental conditions~\cite{molina2017detection, armesto2011inspection}. Inaccurate quality monitoring compromises product integrity, driving up production costs and reducing customer satisfaction~\cite{stavropoulos2023robust}.
Locating weld seams in workpiece imagery is the critical first step; automating this process is essential for downstream defect detection and automated robotic inspection~\cite{kumar2023semi}.
From a computer vision standpoint, welded metallic surfaces are a
challenging domain. Brushed and galvanized steel can produce strong specular reflections that complicate weld imaging; welds can also present irregular and variable geometry; and real shop-floor imagery adds background disturbance, reflections, shadows, blur, and viewpoint variation~\cite{wang2022line,xin2025rapid}.
These factors degrade the performance of deep learning segmentation models in industrial inspection, especially under small annotated datasets, since collecting and labeling industrial images is costly and time-consuming~\cite{tabernik2020segmentation}.
The goal of this work is to assess the feasibility of automatic weld seam \emph{instance segmentation} across acquisition scenarios and imaging modalities, and to identify the factors that dominate performance in practice. Specifically, we compare: (i) RGB images acquired in a controlled laboratory setup versus handheld acquisition in an uncontrolled shop-floor environment;
(ii) conventional RGB imaging versus polarimetric imaging, from which six
per-pixel maps (Intensity, AoLP, DoLP, $I_{\max}$, $I_{\min}$, Specular) are derived;
(iii) segmentation architectures spanning two families (single-stage CNNs
and transformers) and more than an order of magnitude in parameter count; and
(iv) training factors including input resolution, weight initialization, optimizer, and data augmentation. All experiments follow a unified protocol with data partitions defined at the level of the physically acquired weld sample.

\section{Literature review}
\label{sec:related}

Literature on automated weld inspection spans five core themes: seam segmentation, architecture selection, polarization imaging, small-data learning, and deployment efficiency. While lightweight vision pipelines dominate practical implementations, evidence for polarimetric weld segmentation remains sparse, drawing largely on general reflective-surface inspection~\cite{yang2024cvt,goodarzi2022comparison,nandagopal2025robotic}.

\subsection{Vision-based weld inspection}
Although vision-based weld inspection has progressed from classical edge- and morphology-based pipelines toward learned segmentation, classical methods persist in robotic welding for their interpretability and ease of closed-loop integration~\cite{papavasileiou2025quality}. Manual inspection remains common in robotic quality control due to subtle defect appearances, long training times, and variable inspection standards~\cite{nandagopal2025robotic,papavasileiou2025quality}.
Recent deep learning research focuses on pixel-level seam segmentation capable of handling arc light, spatter, and weak boundaries~\cite{yang2024cvt}. A prominent example is CvT-UNet, which uses a U-shaped convolution-transformer encoder--decoder to reach mean IoUs up to 93.75\% across varied environments~\cite{yang2024cvt}.
DSNet targets the real-time regime, reporting 78.01\% IoU, 87.64\% Dice (the overlap-based F1 score on the predicted masks) and 100
FPS~\cite{chen2024dsnet}.

\subsection{Segmentation architectures}
Instance segmentation has historically been dominated by
detection-then-segmentation designs such as Mask R-CNN~\cite{he2017mask}, while newer query-based methods enable direct end-to-end prediction.
Mask2Former serves as a primary baseline, framing semantic, instance, and panoptic segmentation through a shared masked-attention decoder~\cite{cheng2022masked}, though it relies on heavy pixel decoders and has not established clear superiority on real-time benchmarks~\cite{he2023fastinst}. FastInst mitigates this computational bottleneck with lighter decoders, reporting 32.5~FPS and 40.5~AP on COCO~\cite{he2023fastinst}, while DN-DETR accelerates convergence in DETR-style networks via query denoising~\cite{li2022dn}.
For weld inspection, however, YOLO variants remain far more prevalent than DETR-family models. Some segment
weld seams directly, such as an improved YOLOv8s-Seg for seam
tracking~\cite{zhao2024welding} and the boundary-enhanced
SABE-YOLO~\cite{wen2025sabe}; others target detection and ROI localization, including S-YOLO~\cite{wang2025weld}, tailored architectures for seam identification~\cite{qu2025research,xiao2026weld}, and active--passive vision-fusion frameworks~\cite{hu2025weld}. These implementations uniformly prioritize low latency, minimal footprint, and edge deployment.

\subsection{Polarimetric imaging}
Polarimetric imaging measures the polarization state of light reflected by a surface, which depends on the surface's geometry, roughness and material and is largely invisible to conventional intensity imaging. This makes it well suited to reflective metallic inspection: it captures surface and material cues that RGB misses, suppressing glare and improving uniformity~\cite{zuo2023chip,yu2025defect}. Recent sensor-level advances, such as chip-integrated full-Stokes imagers, enable single-shot acquisition in CMOS-compatible form factors, and division-of-focal-plane systems reconstruct Stokes parameters with DoLP and AoLP at full resolution~\cite{zuo2023chip,chen2025computational}.
Polarimetric cues further support per-pixel material discrimination~\cite{kurachi2025one} and 3D reconstruction of specular surfaces when fused with deflectometry~\cite{wang20253d}.

Applied industrial evidence is strongest in adjacent defect-inspection domains rather than in weld segmentation itself. A polarization system for highly reflective curved surfaces removed glare and improved image uniformity, with its YOLOv11 detector gaining 3.9\% in precision over the best baseline~\cite{yu2025defect}, while a dual-stream polarization-plus-RGB network for rail defects reached 73.00\% mIoU on low-contrast segmentation~\cite{pan2025enhanced}. However, direct evaluation of polarimetric imaging for weld-seam instance segmentation remains largely unexplored.

\subsection{Deep learning with small industrial datasets}
Small and imbalanced industrial datasets remain a primary constraint in applied deep learning, since collecting defective samples is expensive and limited diversity undermines robustness~\cite{liu2021defect,dai2022deep}.
Transfer learning from natural-image pretraining is the standard mitigation strategy.
On injection-molding defects, model-based transfer learning with
augmentation reached about 99\% accuracy from only 200 images per class, against 88.7\% for conventional CNNs~\cite{liu2021defect}; on small fused-deposition-modeling (FDM), ImageNet fine-tuning with augmentation exceeded 90\% accuracy on low-cost hardware~\cite{kim2022systematic}. Welding shares these
constraints, with severe class imbalance motivating GAN-based augmentation and pretrained backbones~\cite{dai2022deep}.
Cross-domain evidence further warns that larger models do not automatically transfer better, and that random validation splits can overestimate deployment quality~\cite{goodarzi2022comparison}. Consequently, the trade-off between model capacity and generalization is critical for small weld datasets, where nano-scale CNNs may match larger transformers that lack the data for stable optimization~\cite{yang2024cvt,li2022dn}.

\subsection{Real-time deployment}
For robotic and workstation-based inspection, the literature consistently prioritizes computational efficiency alongside segmentation accuracy rather than peak benchmark scores~\cite{lu2026dsgnet,soori2024intelligent}; DSGNet, for instance, reaches competitive surface-defect mIoU with only 0.49M parameters~\cite{lu2026dsgnet}. This requirement is particularly pronounced in weld inspection, where high headline scores are typically obtained under controlled acquisition and for task-specific formulations: an improved YOLOv8s-Seg seam-tracking robot reports a weld-recognition mAP\textsubscript{50} of about 97.8\% with a 4.88 MB model and roughly 54 ms inference on a Jetson Nano~\cite{zhao2024welding}, while SABE-YOLO attains 127 FPS with 6.6M parameters~\cite{wen2025sabe}. Such figures, obtained in favorable and often single-task settings, leave open how segmentation behaves when acquisition is uncontrolled or the viewpoint changes at deployment time, precisely the regime this study targets. System-level studies reinforce that segmentation quality alone is insufficient: path planning, viewpoint placement and workflow integration all matter in practice~\cite{nandagopal2025robotic,soori2024intelligent}. Few studies, however, evaluate weld segmentation on partner-owned industrial data under both controlled and uncontrolled acquisition~\cite{he2023fastinst,nandagopal2025robotic}.

\subsection{Research gap and contribution}
While existing literature demonstrates the viability of deep weld
segmentation, the benefits of polarization for reflective surface, and the need for lightweight models, it lacks a unified evaluation comparing RGB and polarimetric imaging under matched protocols~\cite{pan2025enhanced,yu2025defect,he2023fastinst}. Furthermore, current studies rarely isolate the impact of acquisition setup relative to architecture choice when operating on actual shop-floor data, despite evidence that real-world domain shift heavily influences model behaviour~\cite{goodarzi2022comparison}. Finally, despite the strong performance of transformer-based architectures on standard benchmarks, there is little empirical evidence regarding their transferability to small industrial weld datasets, leaving it unclear whether higher-capacity models offer any practical benefit over compact CNN alternatives in this regime~\cite{cheng2022masked,he2023fastinst,liu2021defect}.
Building on these gaps, this study delivers three key findings. We first present a unified, threshold-independent evaluation comparing RGB and polarimetric imaging for weld seam instance segmentation, testing both convolutional and transformer architectures on identical dataset splits and evaluation protocols. Next, we measure the impact of acquisition conditions relative to model selection. The data demonstrates that transitioning from controlled to uncontrolled environments causes a performance degradation that outweighs the variances between model families, while multi-map polarimetric imaging effectively restores robustness on unmanaged surfaces without physical illumination controls. Finally, we analyze model behavior under deployment-time viewpoint shifts, observing that detection transformers generalize to unseen, close-range perspectives significantly better than the evaluated CNNs, an effect confirmed to be independent of input resolution and evaluation threshold parameters.
Rather than searching for an optimal model architecture, this work serves as a controlled feasibility study. The goal is to isolate how acquisition modality, illumination control, and architectural choices interact under realistic industrial constraints. Accordingly, quantitative performance figures are presented to evaluate these trade-offs rather than to assert a single state-of-the-art result.

\section{Materials and methods}
\label{sec:method}

\begin{figure*}[htbp]
\centering
\includegraphics[width=\textwidth,height=0.92\textheight,keepaspectratio]{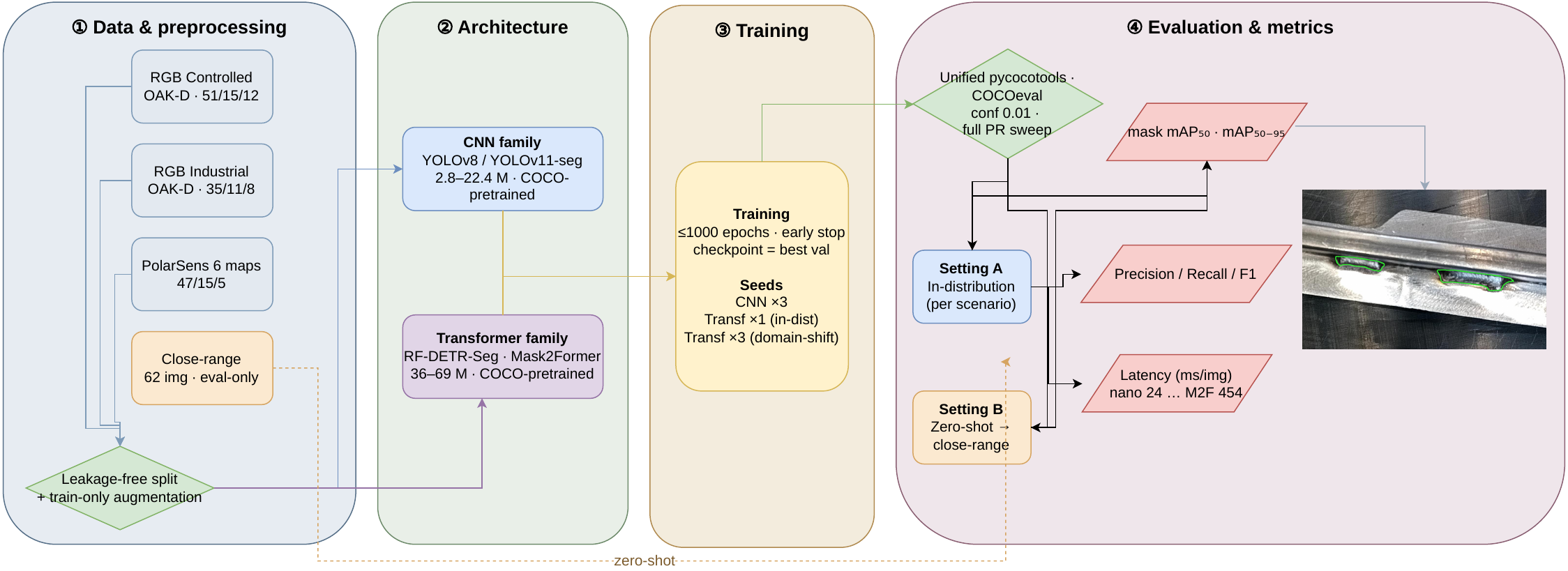}
\caption{End-to-end validation pipeline. Data partitions are split at the
physical-weld level before any augmentation (Sec.~\ref{sec:protocol}); all
six models are evaluated with the same \texttt{pycocotools} protocol at a
fixed confidence threshold (Sec.~\ref{sec:protocol}), both in-distribution
and, zero-shot, on the close-range set (Sec.~\ref{sec:discussion}).}
\label{fig:pipeline}
\end{figure*}

Figure~\ref{fig:pipeline} summarizes the end-to-end validation pipeline
followed in this study, from data acquisition to the final metrics; the
remainder of this section details each stage.

\subsection{Materials}
\label{sec:datasets}

All images depict weld seams on metallic specimens provided by the industrial partner specialized in custom operator cabin manufacturing. Ground-truth annotations are polygonal instance masks of a single class (\emph{Weld}), produced manually. Table~\ref{tab:datasets} summarizes the datasets.

\begin{table*}[t]
\centering
\footnotesize
\caption{Datasets used in this study. ``Train'' counts physically distinct
source acquisitions, with augmented training images in parentheses;
augmented variants are generated from training sources only
(Sec.~\ref{sec:protocol}). Acquisition conditions are detailed in the text.
The close-range set is evaluation-only: by design it has no train or
validation split (Sec.~\ref{sec:datasets}, ``Close-range test set''), and all
62 images are used exclusively for zero-shot cross-evaluation of models
trained on the other scenarios.}
\label{tab:datasets}
\vspace{6pt}
\begin{tabular}{llcrrr}
\toprule
Dataset & Camera & Resolution & Train (aug.) & Val & Test \\
\midrule
RGB controlled      & OAK-D     & $1920{\times}1080$ & 51 (510)  & 15 & 12 \\
RGB industrial      & OAK-D     & $1920{\times}1080$ & 35 (175)  & 11 & 8  \\
PolarSens multi-map & PolarSens & $2448{\times}2048$ & 47 (1384) & 15 & 5  \\
PolarSens Intensity & PolarSens & $1224{\times}1024$ & 45 (450)  & 14 & 10 \\
Close-range test    & OAK-D     & $1920{\times}1080$ & ---       & --- & 62 \\
\bottomrule
\end{tabular}
\end{table*}

\paragraph{Controlled RGB dataset} Images were acquired with an OAK-D RGB camera ~\cite{luxonis_oakd}  fixed at 20\,cm from the workpiece, with two auxiliary polarizer-equipped spotlights and a neutral black background, minimizing environmental variability. This setup provides an upper-bound scenario for model performance under consistent acquisition. Three metal specimens were
imaged, for a total of 78 source images.

\paragraph{Industrial RGB dataset} A second set of 54 images was captured directly at the industrial production facility under unconstrained conditions. These acquisitions lack background or illumination control and include specular reflections, cluttered backgrounds, variable perspective angles, and non-uniform lighting to reflect operational deployment conditions.

\begin{figure*}[t]
\centering
\includegraphics[width=\linewidth]{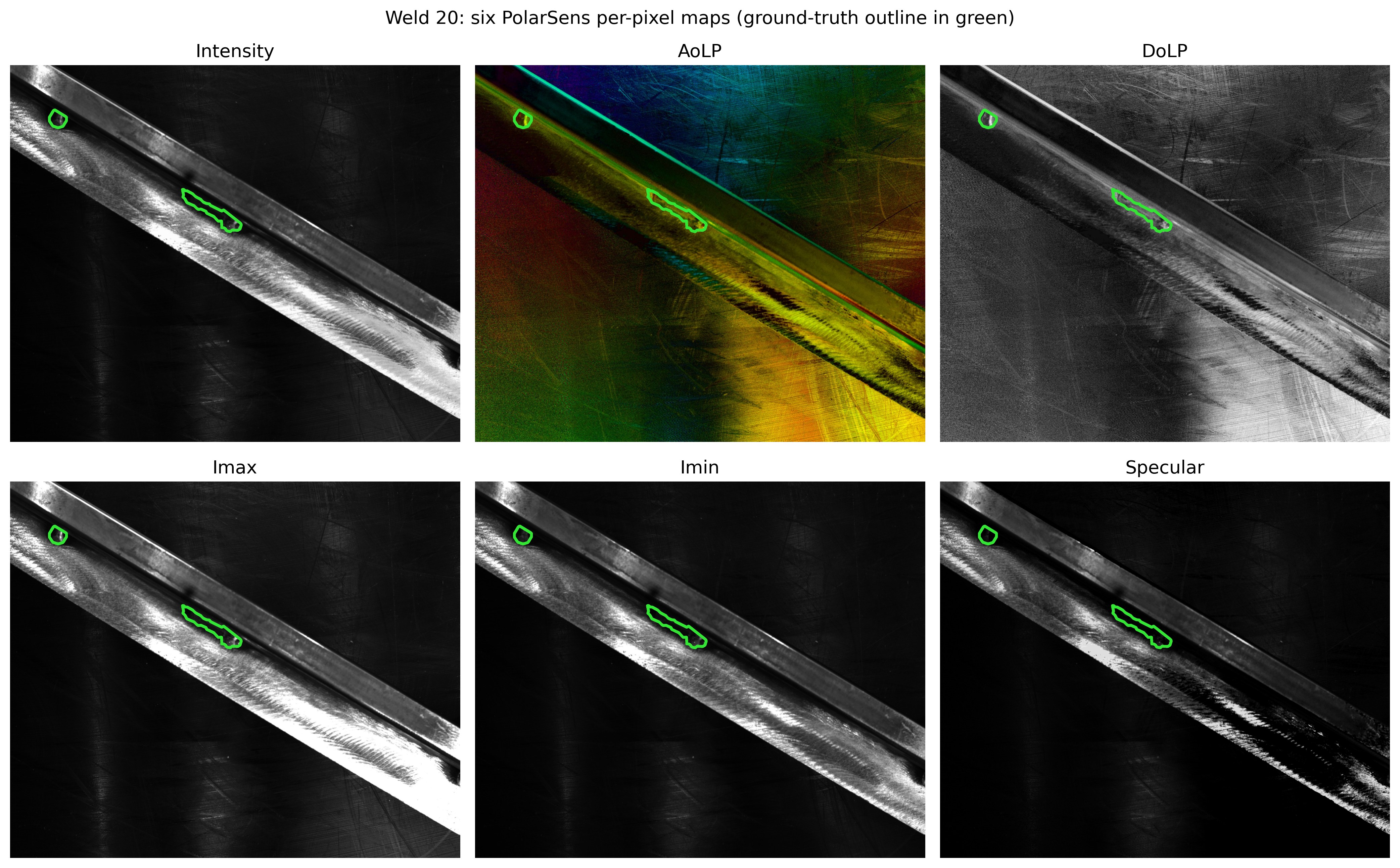}
\caption{The six PolarSens per-pixel maps derived from one acquisition
(weld 20, test split), ground-truth weld outline in green.}
\label{fig:polar-maps}
\end{figure*}

\paragraph{PolarSens datasets} A division-of-focal-plane polarization camera (Baumer PolarSens, $0^\circ/45^\circ/90^\circ/135^\circ$ micro-polarizer array) was used to acquire 67 physically distinct weld scenes under polarized illumination. From each acquisition, six per-pixel maps are derived, namely Intensity, angle of linear polarization (AoLP), degree of linear polarization (DoLP), $I_{\max}$, $I_{\min}$ and Specular, each stored as a 3-channel image. These maps carry complementary information on surface structure, orientation and roughness, and allow assessing the contribution of polarization cues on reflective metallic surfaces. Because the polarization camera has a different viewpoint, this dataset is not
pixel-aligned with the RGB ones.
Figure~\ref{fig:polar-maps} illustrates an example of the six polarization maps for a representative weld seam. Unlike the five magnitude-based maps (intensity, DoLP, $I_{\max}$, $I_{\min}$, and specular), which present similar grayscale appearances with distinct highlight distributions, the AoLP map exhibits a characteristic hue-like angular pattern. This spatial variance across representations provides the rationale for the multi-map strategy detailed in Sec.~\ref{sec:polstrategies}.

\paragraph{Close-range test set} An additional evaluation-only dataset of 62 close-range images ($\sim 10\,\text{cm}$, single weld per image, manually captured) evaluates model generalization to the acquisition geometry expected in a robot-mounted camera setup.

\subsection{Segmentation models}
\label{sec:models}
We compare six instance segmentation models spanning two architectural families and more than an order of magnitude in parameter count (Table~\ref{tab:models}): four single-stage CNN detectors with mask heads from the YOLO family (YOLOv11-n/s/m-seg~\cite{ultralytics_yolov11, khanam2024yolov11}, YOLOv8-s-seg~\cite{ultralytics_yolov8}), and two transformer-based models: RF-DETR-Seg (Small)~\cite{roboflow_rfdetr}, a DINOv2-backbone detection transformer with a segmentation head, and Mask2Former (Swin-Small) with masked-attention segmentation ~\cite{cheng2022masked}. This models selection balances inference throughput and capacity for industrial deployment. Transformer architectures are included to assess their transferability to data-scarce manufacturing settings and to anchor the upper bound of model capacity within our benchmark. All architectures are fine-tuned from COCO-pretrained weights.

\begin{table*}[t]
\centering
\footnotesize
\setlength{\tabcolsep}{4pt}
\caption{Evaluated architectures. End-to-end latency (preprocessing,
inference, postprocessing) per image on a single NVIDIA RTX 3090, measured
on the polarimetric validation split at each model's input resolution;
transformer figures include amortized model loading.}
\label{tab:models}
\vspace{6pt}
\begin{tabular}{llrrl}
\toprule
Model & Family & Params (M) & Input res. (px) & Latency (ms/img) \\
\midrule
YOLOv11-n-seg & CNN & 2.8 & 2464 & 24 \\
YOLOv11-s-seg & CNN & 10.1 & 2464 & 49 \\
YOLOv11-m-seg & CNN & 22.4 & 2464 & 111 \\
YOLOv8-s-seg  & CNN & 11.8 & 2464 & 42 \\
RF-DETR-Seg-S & Transformer & 36.3 & 1120 & 132 \\
Mask2Former (Swin-S) & Transformer & 69 & 1024$^{\dagger}$ & 454 \\
\bottomrule
\end{tabular}

\smallskip
{\footnotesize $^{\dagger}$shorter side; transformer models are trained at
reduced resolution due to GPU memory constraints, a caveat discussed in
Sec.~\ref{sec:discussion}.}
\end{table*}

\subsection{Training strategies}
\label{sec:training}
All models are trained to convergence, with a maximum budget of 1000 epochs
and early stopping. To train the CNN models, we use the Ultralytics
pipeline~\cite{ultralytics_yolov11} with its default hyperparameters: the SGD
optimizer with an initial learning rate $lr_0{=}0.01$ and a batch size of 2 at
full resolution. The transformer models are trained with the AdamW
optimizer~\cite{loshchilov2019decoupledweightdecayregularization} within their respective reference
implementations: a learning rate of $5{\times}10^{-5}$ for Mask2Former, and
the RF-DETR defaults with an effective batch size of 16 obtained through
gradient accumulation. All experiments run on a single NVIDIA RTX 3090
(24\,GB). Wall-clock training time scales with both model capacity and
dataset size (largest for the augmented PolarSens multi-map set, smallest
for industrial RGB): across scenarios and seeds, CNN training took on
average 1.5\,h (YOLOv11-n, range 0.2--7.6\,h), 1.8\,h (YOLOv11-s, range
0.3--6.2\,h), 4.7\,h (YOLOv11-m, range 0.8--15.6\,h) and 1.6\,h (YOLOv8-s,
range 0.3--4.4\,h); the transformers took 0.7--5.5\,h (RF-DETR-Seg-S) and
0.9--9.6\,h (Mask2Former), the same industrial-RGB-to-PolarSens range as
the CNNs but consistently at the upper end of it, consistent with their
larger parameter counts (Table~\ref{tab:models}).
For the RGB scenarios, we additionally compare pretrained versus
from-scratch initialization; native ($1920{\times}1080$) versus halved
($960{\times}540$) input resolution; and, for the best configurations, a
hyperparameter-optimization stage comparing the SGD and AdamW optimizers,
using Ultralytics' built-in genetic-algorithm tuner. Each ``iteration'' is
one hyperparameter trial: it trains an independent model for 50 epochs with
its own sampled hyperparameter values, and only the best-performing trial's
hyperparameters are carried forward to the final model. The search spends
on the order of $100\times50=5000$ epochs of training compute in total,
split across 100 short, independent 50-epoch runs.
Some datasets are augmented offline: augmented image variants are generated
once from the training images and stored on disk, rather than being produced
on the fly during training (the procedure is detailed in
Sec.~\ref{sec:protocol}). For these pre-augmented datasets, the online
augmentation of the training pipeline is disabled, so that all image-level
variability comes from the stored offline augmentation and is identical
across the models being compared.

\paragraph{Seed repetition}
Reported metrics are mask mAP (mean average precision computed on the
predicted segmentation masks, as opposed to bounding boxes). To separate
genuine architectural effects from run-to-run randomness on our small test
sets, each of the four CNN configurations is trained with three random seeds
(varying weight initialization and data shuffling), and we report the mean
and standard deviation across these seeds throughout Sec.~\ref{sec:results}.
Training the transformer models is substantially more expensive, so their
in-distribution results (Tables~\ref{tab:rgb-lab}, \ref{tab:rgb-ind},
\ref{tab:polar-multimap}) are obtained with a single seed. The practical
consequence is that small in-distribution gaps involving the transformers
should be read with caution: for example, RF-DETR's in-distribution lead over
the best CNN on industrial RGB (Sec.~\ref{sec:res-rgb-ind}) is only about
three to four times the CNN seed standard deviation, so we cannot yet rule out
that part of it is seed noise. Confirming these smaller in-distribution
margins would require training the transformers with multiple seeds as well,
which we leave to future work.
The one comparison for which we train the transformers with three seeds is the domain-shift cross-evaluation to the close-range set (Sec.~\ref{sec:discussion}), where both transformers are retrained with three seeds for each source scenario (controlled and industrial RGB). Here the effect is an order of magnitude larger than the CNN seed variance, as the transformer-versus-CNN gap in mask mAP\textsubscript{50} spans 0.6 to 0.8, compared to a maximum CNN seed standard deviation of 0.05. The multi-seed cost is therefore justified precisely where the claim is strongest. These same three
industrial-RGB seeds are additionally re-evaluated in-distribution under the
identical protocol, to check that RF-DETR's out-of-distribution advantage
reproduces across runs rather than resting on one lucky seed; this does not
extend multi-seed training to the general in-distribution rankings of
Sec.~\ref{sec:res-rgb-ind}.

\subsection{Polarimetric strategies}
\label{sec:polstrategies}
\begin{figure*}[htbp]
\centering
\includegraphics[width=\textwidth]{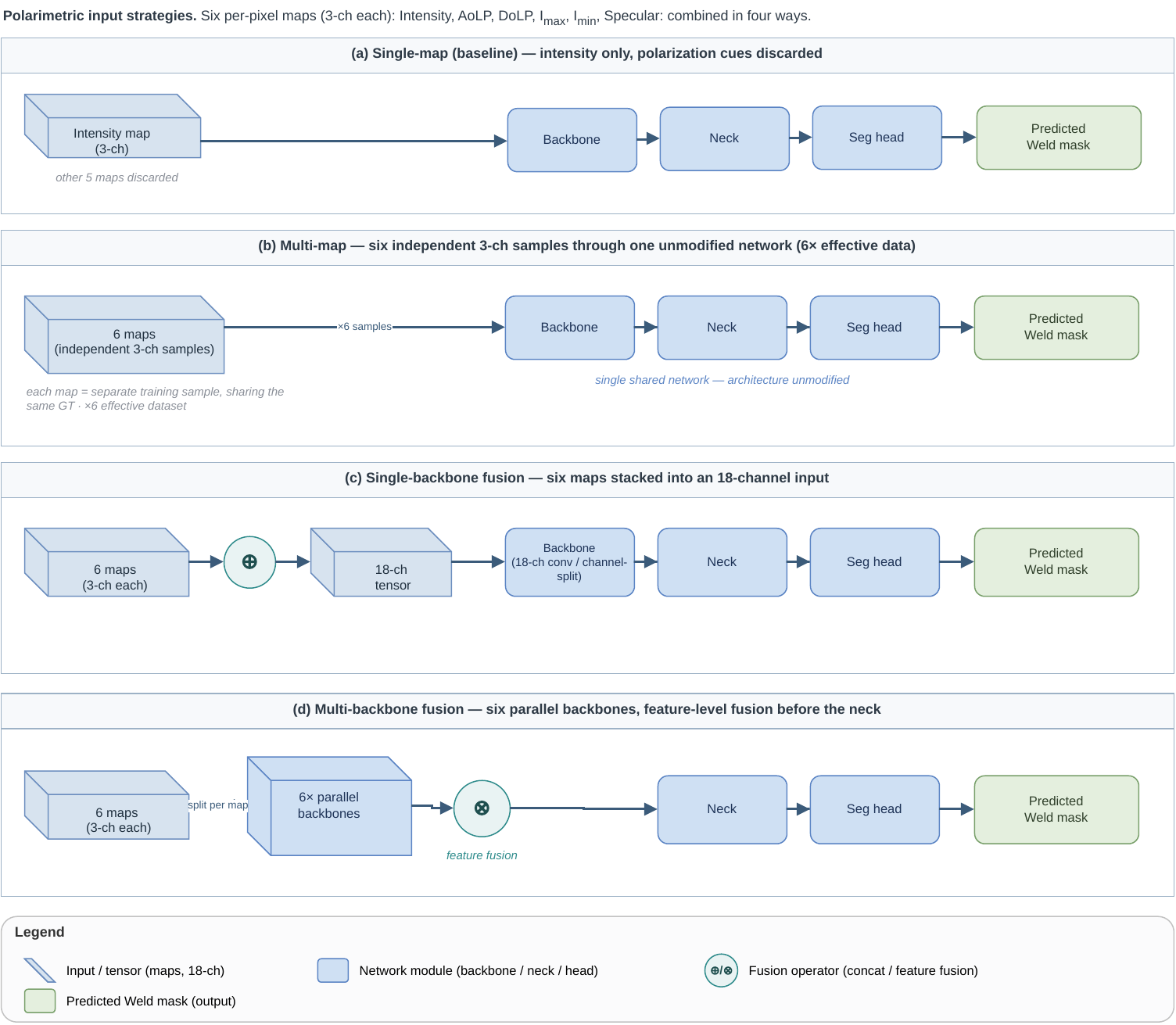}
\caption{The four polarimetric strategies of Sec.~\ref{sec:polstrategies}.
(a)--(b) use an unmodified YOLO11-seg architecture, single 3-channel input;
they differ only in what is fed as training data. (c)--(d) use custom
\texttt{Split} / \texttt{ConvMultiInput} layers to fuse all six maps inside
the network, either through one shared backbone on the stacked 18-channel
input (c) or through six parallel backbones concatenated at three feature
scales before a shared neck (d). Verified against
\texttt{config/yolo11-seg-custom-polarsens-\{singlebackbone,multibackbone\}.yaml}.}
\label{fig:polar-arch}
\end{figure*}

We evaluate four strategies to leverage the polarimetric information (Figure~\ref{fig:polar-arch}):
\begin{enumerate}
  \item \textit{Single-map} (baseline): only the intensity map is used, as a
        standard 3-channel input, discarding the polarization cues.
  \item \textit{Multi-map}: all six maps are treated as independent 3-channel
        training samples that share the same ground-truth annotation, leaving
        the network architecture unmodified. In practice this multiplies the
        effective training-set size sixfold, since each map of a given weld is
        added to the dataset as a separate image with the same mask.
  \item \textit{Single-backbone fusion}: the six maps are stacked into an
        18-channel input processed by a single backbone, evaluated with both a
        direct 18-channel first-layer convolution and an explicit
        channel-split layer.
  \item \textit{Multi-backbone fusion}: six parallel backbones, one per map,
        with feature-level fusion preceding the neck.
\end{enumerate}
For the two fusion strategies (3 and 4), data augmentation must strictly
preserve spatial alignment across all six maps. Geometric transformations
(horizontal/vertical flips, rotations within $\pm15^\circ$, and shears within
$\pm10^\circ$) are therefore sampled once per sample and applied identically
to all six maps and to the polygon masks. Photometric augmentations are
intentionally omitted, since they would corrupt the physical meaning of the
polarimetric quantities. Results for strategies 3 and 4 are presented in
Sec.~\ref{sec:res-fusion}.

\subsection{Experimental protocol}
\label{sec:protocol}

Data partitions are strictly structured at the physical acquisition level: all original images and augmented variants of a given weld sample reside within the same partition to prevent data leakage. Augmentation is restricted exclusively to the training set, while validation and test sets consist solely of unaugmented original images. Within each experimental scenario, all models are trained and evaluated on identical splits. Evaluation metrics comprise COCO-style box and mask mAP\textsubscript{50} and mAP\textsubscript{50--95}, alongside precision, recall, and F1 score at the selected operating point.
To ensure a fair comparison across architectural paradigms, all CNN and transformer models are evaluated using a unified \texttt{pycocotools} pipeline~\cite{cocoapi} at a fixed low confidence threshold ($0.01$). This allows \texttt{COCOeval}~\cite{lin2014microsoft} to sweep confidence thresholds across the full precision--recall spectrum. This threshold adjustment accounts for the lower calibrated confidence scores of query-based transformers. Applying higher operational thresholds (e.g., $0.5$) truncates valid predictions and biases average precision against CNN baselines.
The validation split plays a single, specific role: it is the signal used to pick the checkpoint that is later evaluated, for every architecture in this study. Concretely, the CNN training pipeline retains the checkpoint with the best validation fitness at each epoch~\cite{ultralytics_yolov11}, Mask2Former retains the checkpoint with the lowest validation loss, and RF-DETR retains the checkpoint with the best validation mAP\textsubscript{50-95}. The test split is never seen during training or checkpoint selection; it is touched only for the final, single evaluation reported in each table. Checkpoint selection optimizes a validation-set criterion, so validation scores could plausibly run systematically higher than test scores; empirically they don't, in this study, in either direction. We verified that validation and test partitions never share a physical weld or an augmented variant of one, across every scenario, ruling out a split-leakage bug. Two factors explain the pattern instead. The validation and test splits are both small (8--15 and 5--12 images respectively, per scenario), so ranking noise from a handful of images is comparable in size to any genuine gap between splits. And the checkpoint-selection criterion (validation fitness, validation loss, or validation mAP\textsubscript{50-95}, depending on the framework) differs from the mask mAP\textsubscript{50} column reported in the tables, so the checkpoint favored by the selection criterion can score differently on that specific metric. Results tables report both test and validation for every model so this can be inspected directly.
Within each results table, models are listed in descending order of test-split mask mAP\textsubscript{50}, independently in every table; this produces a different row order in different tables and is not a stable per-model ranking carried over from one scenario to the next, which is the point being illustrated (Sec.~\ref{sec:res-arch}).

\section{Results}
\label{sec:results}

\begin{figure*}[t]
\centering
\includegraphics[width=\linewidth]{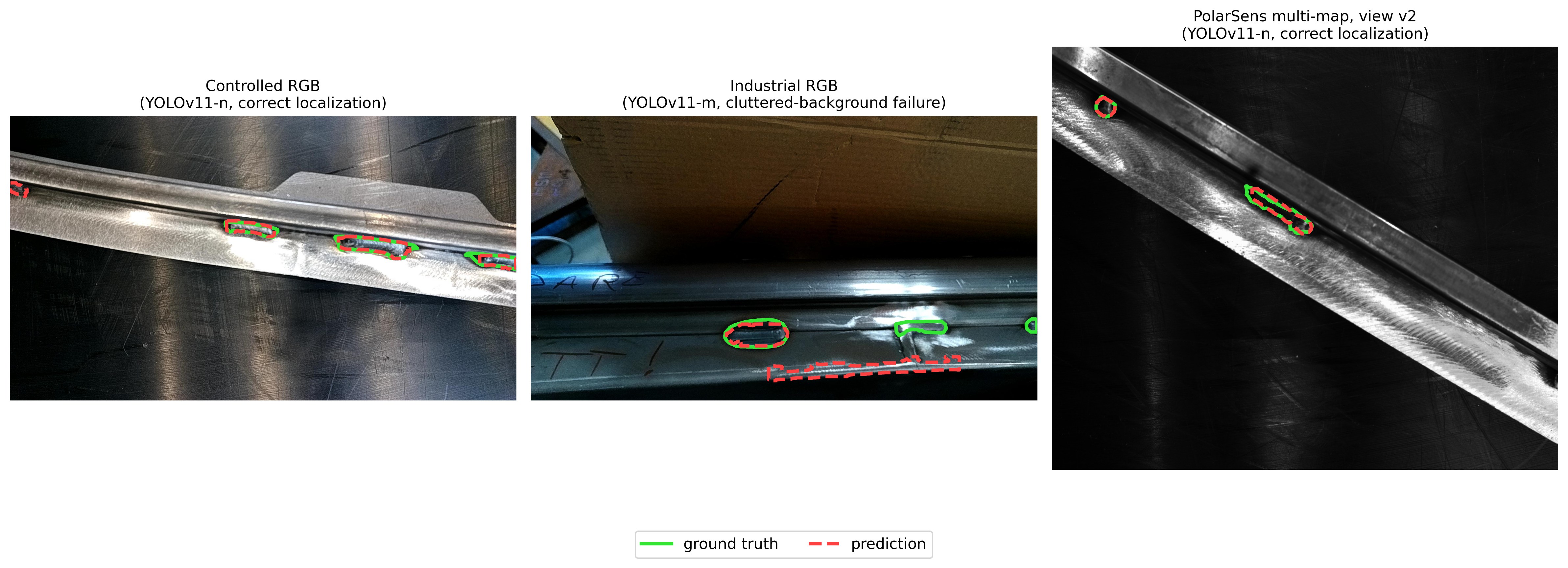}
\caption{Qualitative segmentation results (ground truth in solid green,
prediction in dashed red) across the three main acquisition scenarios.
Industrial RGB shows a representative CNN failure mode: a false-positive
mask on cluttered background (bottom) alongside a missed weld (top right,
green outline with no matching prediction).}
\label{fig:qualitative}
\end{figure*}

Figure~\ref{fig:qualitative} previews the three acquisition scenarios qualitatively before the detailed quantitative evaluation: correct segmentation in controlled RGB and PolarSens maps, and a representative CNN failure mode under cluttered industrial RGB (false-positive background clutter and a missed weld), providing a concrete example of the limitations detailed in Sec.~\ref{sec:res-rgb-ind}.

\subsection{Controlled RGB}
\label{sec:res-rgb-lab}

Table~\ref{tab:rgb-lab} reports model performance under controlled conditions using the uniform \texttt{pycocotools} evaluation protocol (Sec.~\ref{sec:protocol}) for seed 0.
On the test split, RF-DETR-Seg achieves the highest performance (mask mAP\textsubscript{50} of $0.900$), though it converges to the CNN cluster on the validation set ($0.755$, compared to $0.72\text{--}0.81$ for CNNs). Mask2Former lags across both splits ($0.61\text{--}0.69$). The four CNN architectures cluster tightly on both splits (test 0.79--0.82),
with no consistent ranking by size. To quantify variance within the CNN family, Table~\ref{tab:seed-variance} provides multi-seed statistics across three random runs, evaluated under the Ultralytics-native validation protocol rather than the uniform \texttt{pycocotools} protocol used in Table~\ref{tab:rgb-lab}: the two protocols score the same checkpoint differently, so for a given architecture the seed-0 entry in Table~\ref{tab:rgb-lab} and the seed-0 sample underlying Table~\ref{tab:seed-variance} are not the same number, only the same trained model. In the controlled RGB rows of Table~\ref{tab:seed-variance}, every CNN architecture yields a three-seed mean mask mAP\textsubscript{50--95} between $0.46$ and $0.55$, with standard deviations ranging from $0.005$ to $0.026$. The YOLOv11 variants (nano, small, medium) fall within half a standard deviation of one another, with YOLOv8-s showing the largest deviation ($\sim 1.6$ pooled standard deviations on test). Given the small test partitions (8--15 images per split) and three-seed sampling, these minor variations remain within expected noise bounds, preventing a definitive ranking among CNN backbones. Earlier feasibility trials showed that dedicated hyperparameter optimization (100 SGD tuning iterations) boosted YOLOv11-s/m mask mAP\textsubscript{50} to $0.89\text{--}0.90$ (and up to $0.85$ on the close-range set). However, achieving these gains over the default baselines requires significant tuning overhead.
Direct zero-shot evaluation on the close-range set yields a distinct trend (Table~\ref{tab:closerange-lab}). Averaged over three seeds, RF-DETR-Seg generalizes best to the new domain (mAP\textsubscript{50} = 0.842, mAP\textsubscript{50--95} = 0.537), outperforming CNN baselines ($0.08\text{--}0.17$) and Mask2Former ($0.41$) by a wide margin. In Sec.~\ref{sec:discussion}, we analyze these results and run a resolution-matched ablation to isolate architectural effects from image resolution.

\begin{table*}[t]
\centering
\footnotesize
\caption{Controlled RGB scenario, uniform \texttt{pycocotools} protocol, seed-0
checkpoints (test: 12 images; validation: 15 images).}
\label{tab:rgb-lab}
\vspace{6pt}
\begin{tabular}{lcccc}
\toprule
& \multicolumn{2}{c}{Test} & \multicolumn{2}{c}{Validation} \\
\cmidrule(lr){2-3}\cmidrule(lr){4-5}
Model & mask mAP\textsubscript{50} & mAP\textsubscript{50--95}
      & mask mAP\textsubscript{50} & mAP\textsubscript{50--95} \\
\midrule
RF-DETR-Seg-S & \textbf{0.900} & 0.471 & 0.755 & 0.395 \\
YOLOv11-n & 0.819 & 0.452 & 0.792 & \textbf{0.436} \\
YOLOv11-s & 0.797 & \textbf{0.475} & \textbf{0.814} & 0.433 \\
YOLOv11-m & 0.795 & 0.437 & 0.783 & 0.433 \\
YOLOv8-s  & 0.791 & 0.454 & 0.723 & 0.422 \\
Mask2Former (Swin-S) & 0.610 & 0.355 & 0.686 & 0.361 \\
\bottomrule
\end{tabular}
\end{table*}

\begin{table}[htbp]
\centering
\footnotesize
\caption{Zero-shot cross-evaluation on the close-range test set (62 images)
of the six models trained on controlled RGB (Table~\ref{tab:rgb-lab}),
uniform \texttt{pycocotools} protocol; no fine-tuning on close-range data.
Mean $\pm$ population standard deviation over three training seeds for every
model. Even the worst individual Mask2Former seed (0.357) exceeds the best
individual CNN seed (YOLOv11-s, 0.215), so the transformer advantage is not
an artifact of seed selection.}
\label{tab:closerange-lab}
\vspace{6pt}
\begin{tabular}{lcc}
\toprule
Model & mask mAP\textsubscript{50} & mAP\textsubscript{50--95} \\
\midrule
RF-DETR-Seg-S         & $\mathbf{0.842 \pm 0.015}$ & $\mathbf{0.537 \pm 0.027}$ \\
Mask2Former (Swin-S)  & $0.406 \pm 0.050$ & $0.243 \pm 0.042$ \\
YOLOv11-s             & $0.168 \pm 0.037$ & $0.091 \pm 0.019$ \\
YOLOv11-n             & $0.116 \pm 0.005$ & $0.066 \pm 0.004$ \\
YOLOv11-m             & $0.102 \pm 0.011$ & $0.048 \pm 0.008$ \\
YOLOv8-s              & $0.081 \pm 0.005$ & $0.045 \pm 0.004$ \\
\bottomrule
\end{tabular}
\end{table}

\subsection{Industrial RGB}
\label{sec:res-rgb-ind}

Under uncontrolled shop-floor acquisition, performance drops sharply for the four CNN architectures (Table~\ref{tab:rgb-ind}): on the seed-0 checkpoints, the best CNN reaches test mask mAP\textsubscript{50} 0.47 (YOLOv8-s, mAP\textsubscript{50--95} 0.24), against 0.79--0.82 for the same four models in controlled conditions (Table~\ref{tab:rgb-lab}), a performance drop that far exceeds any difference among the CNN variants. Table~\ref{tab:seed-variance} confirms this with three-seed statistics: CNN test means cluster at 0.22--0.48 mask mAP\textsubscript{50}, with standard deviations of 0.04--0.05; under this multi-seed average, YOLOv11-n emerges as the top-performing CNN (mean 0.477) rather than YOLOv8-s (mean 0.440), which led under seed 0 alone. This shift underscores the instability of single-seed rankings. Across runs, only YOLOv11-m consistently underperforms beyond seed variance, whereas the relative rankings among the nano, small, and YOLOv8-s variants remain statistically indistinguishable. RF-DETR is the best model overall on both splits (test 0.619/0.243, validation 0.643/0.311),
clearly ahead of every CNN. Mask2Former is second on test (0.523/0.198) but drops into the CNN cluster on validation (0.435/0.212, below three of the four CNNs). This is the only in-distribution scenario in which a transformer outperforms every CNN outright.
Zero-shot cross-evaluation on the close-range set (Table~\ref{tab:closerange-ind}) provides the clearest evidence of this trend across all model comparisons. When both transformers are fine-tuned on the industrial RGB images and then
evaluated, without any further training, on the close-range set, they
outperform every CNN baseline by a large margin under three-seed statistics.
RF-DETR reaches a three-seed mean mask mAP\textsubscript{50} of $0.810$
(mAP\textsubscript{50--95} $0.537$) and Mask2Former $0.461$ ($0.300$), whereas
the four CNNs span only $0.043$--$0.189$ ($0.018$--$0.089$). In other words,
the weakest transformer result is still more than twice the strongest CNN
result on this set.
Two caveats temper the Mask2Former figure specifically. Its three-seed mean is
inflated by a single unusually strong seed: the three runs score $0.352$,
$0.338$ and $0.693$, giving a standard deviation ($0.164$) far above the
$\le0.05$ seen everywhere else in this study. Even so, its \emph{worst}
individual seed ($0.338$) still exceeds the \emph{best} individual CNN seed
($0.200$), so the ranking does not depend on that lucky run; the high variance
is a separate finding, revisited in Sec.~\ref{sec:discussion}. RF-DETR, by
contrast, is stable across seeds (std $0.007$), so its advantage rests on firm
ground.
This close-range result mirrors the one obtained when the same models are
instead fine-tuned on controlled RGB (Table~\ref{tab:closerange-lab}): the
transformers' advantage on the shifted, close-range viewpoint therefore holds
regardless of which RGB scenario they were trained on. Because this pattern is
the paper's central finding, it is analyzed in full, together with a
resolution-matched ablation that separates an architectural explanation from a
resolution artifact, in Sec.~\ref{sec:discussion}. On the CNN side, a
typical failure mode on this set is background structure misclassified as weld
in visually cluttered scenes.These findings illustrate the drop from lab conditions to field deployment, underpinning our acquisition recommendations in Sec.~\ref{sec:discussion}. In comparison, the preliminary feasibility study scored mask mAP\textsubscript{50} of 0.43/0.36 for YOLOv11-s and 0.49/0.17 for YOLOv11-m (half vs. native resolution, default hyperparameters), consistent with our CNN baselines.

\begin{table*}[t]
\centering
\footnotesize
\caption{Industrial RGB scenario, uniform \texttt{pycocotools} protocol,
seed-0 checkpoints (test: 8 images; validation: 11 images).}
\label{tab:rgb-ind}
\vspace{6pt}
\begin{tabular}{lcccc}
\toprule
& \multicolumn{2}{c}{Test} & \multicolumn{2}{c}{Validation} \\
\cmidrule(lr){2-3}\cmidrule(lr){4-5}
Model & mask mAP\textsubscript{50} & mAP\textsubscript{50--95}
      & mask mAP\textsubscript{50} & mAP\textsubscript{50--95} \\
\midrule
RF-DETR-Seg-S & \textbf{0.619} & 0.243 & \textbf{0.643} & \textbf{0.311} \\
Mask2Former (Swin-S) & 0.523 & 0.198 & 0.435 & 0.212 \\
YOLOv8-s  & 0.471 & \textbf{0.244} & 0.500 & 0.224 \\
YOLOv11-s & 0.414 & 0.167 & 0.477 & 0.209 \\
YOLOv11-n & 0.408 & 0.130 & 0.507 & 0.251 \\
YOLOv11-m & 0.247 & 0.088 & 0.432 & 0.219 \\
\bottomrule
\end{tabular}
\end{table*}

\begin{table}[htbp]
\centering
\footnotesize
\caption{Zero-shot cross-evaluation on the close-range test set of the six
models trained on industrial RGB (Table~\ref{tab:rgb-ind}), uniform
\texttt{pycocotools} protocol; no fine-tuning on close-range data. Mean $\pm$
population standard deviation over three training seeds for every model.
Mask2Former's variance here is exceptional (seeds 0.352, 0.338, 0.693): one
seed nearly doubles the other two. Even so, its worst individual seed (0.338)
exceeds the best individual CNN seed (YOLOv11-n, 0.200), so the ranking is
unaffected; this high transformer variance is itself discussed in
Sec.~\ref{sec:discussion}.}
\label{tab:closerange-ind}
\vspace{6pt}
\begin{tabular}{lcc}
\toprule
Model & mask mAP\textsubscript{50} & mAP\textsubscript{50--95} \\
\midrule
RF-DETR-Seg-S         & $\mathbf{0.810 \pm 0.007}$ & $\mathbf{0.537 \pm 0.014}$ \\
Mask2Former (Swin-S)  & $0.461 \pm 0.164$ & $0.300 \pm 0.124$ \\
YOLOv11-n             & $0.189 \pm 0.008$ & $0.089 \pm 0.005$ \\
YOLOv11-s             & $0.102 \pm 0.037$ & $0.037 \pm 0.011$ \\
YOLOv8-s              & $0.061 \pm 0.016$ & $0.025 \pm 0.007$ \\
YOLOv11-m             & $0.043 \pm 0.015$ & $0.018 \pm 0.010$ \\
\bottomrule
\end{tabular}
\end{table}

\subsection{PolarSens}
\label{sec:res-polar}

\paragraph{Single map} Using the Intensity map alone yields limited
performance (best: mask mAP\textsubscript{50} 0.40, mAP\textsubscript{50--95}
0.23, YOLOv11-s pretrained; not tabulated, as this baseline is not the focus
of the comparison), below the controlled RGB baseline: the intensity content
of the polarimetric sensor, by itself, does not compensate for the reduced
dataset size.

\paragraph{Multi-map without augmentation} Feeding all six maps as
independent samples to an unmodified YOLOv11 improves results markedly
(mask mAP\textsubscript{50} up to 0.70, mAP\textsubscript{50--95} 0.36),
indicating that the complementary polarization maps carry discriminative
information about the weld region even when processed independently.

\paragraph{Multi-map with geometric augmentation} Adding the
alignment-preserving geometric augmentation of
Sec.~\ref{sec:polstrategies} produces the strongest results of the study
(Table~\ref{tab:polar-multimap}). On welds never seen in training, all
four CNN architectures localize welds reliably (test mask
mAP\textsubscript{50} 0.74--0.91) and RF-DETR-Seg is essentially tied with
the best CNN on this metric using the same seed-0 checkpoints as
Table~\ref{tab:polar-multimap} (0.904 versus 0.914). Against the
three-seed mean of the best CNN (0.926, YOLOv11-n,
Table~\ref{tab:seed-variance}) the tie widens slightly, though still by less
than the seed-to-seed variance discussed below. The stricter
mAP\textsubscript{50--95} suggests otherwise: every CNN baseline ($0.38\text{--}0.50$) outperforms RF-DETR ($0.421$) and Mask2Former ($0.309$), indicating that transformers achieve comparable coarse localization to CNNs, but with less precise boundaries. Table~\ref{tab:seed-variance}
quantifies CNN seed variance directly: three-seed test means span
0.537--0.570 mask mAP\textsubscript{50--95} with standard deviations of
0.012--0.030, i.e.\ the four architectures perform comparably on this metric, though YOLOv11-n has the highest mean on
both splits. The gap between the lenient and strict metrics for the CNNs
themselves indicates that residual errors concentrate in boundary
precision rather than in weld localization.

\paragraph{Transformer baselines} When evaluated under the same threshold-independent protocol as the CNNs, RF-DETR-Seg closes the gap on PolarSens significantly more than raw scores suggest: it matches the
best CNN on mask mAP\textsubscript{50} (Table~\ref{tab:polar-multimap})
and trails only on the strict mAP\textsubscript{50--95} metric, where all
four CNNs maintain a distinct lead (0.38--0.50 versus 0.421). Mask2Former
remains the weakest model on both splits and both metrics (0.52/0.31 test).
Both transformers show early validation-loss saturation during training,
indicative of over-parameterization relative to the
$\sim$50 training welds; this training-curve evidence is independent of evaluation thresholding, offering the strongest support for how model capacity limits performance in low-data regimes. Their
latency (132 and 454\,ms/img respectively, versus 24\,ms/img for YOLOv11-n)
remains a practical deployment overhead regardless of accuracy. Part of the remaining performance gap likely stems from resolution constraints imposed by GPU memory limits (Tab.~\ref{tab:models}); disentangling
resolution from architecture effects for this scenario specifically is
left to future work (Sec.~\ref{sec:discussion} reports a resolution
ablation for the RGB scenarios).

\begin{table*}[t]
\centering
\footnotesize
\caption{PolarSens multi-map with geometric augmentation, uniform
\texttt{pycocotools} protocol, seed-0 checkpoints: leakage-free sample-level
splits (test: 5 unseen welds / 30 images; validation: 15 unseen welds /
90 images).}
\label{tab:polar-multimap}
\vspace{6pt}
\begin{tabular}{lcccc}
\toprule
& \multicolumn{2}{c}{Test} & \multicolumn{2}{c}{Validation} \\
\cmidrule(lr){2-3}\cmidrule(lr){4-5}
Model & mask mAP\textsubscript{50} & mAP\textsubscript{50--95}
      & mask mAP\textsubscript{50} & mAP\textsubscript{50--95} \\
\midrule
YOLOv11-n & \textbf{0.914} & \textbf{0.501} & \textbf{0.756} & \textbf{0.350} \\
RF-DETR-Seg-S & 0.904 & 0.421 & 0.701 & 0.295 \\
YOLOv8-s  & 0.862 & 0.500 & 0.700 & 0.336 \\
YOLOv11-s & 0.807 & 0.479 & 0.679 & 0.323 \\
YOLOv11-m & 0.743 & 0.382 & 0.644 & 0.289 \\
Mask2Former (Swin-S) & 0.519 & 0.309 & 0.595 & 0.278 \\
\bottomrule
\end{tabular}
\end{table*}

\begin{table*}[t]
\centering
\footnotesize
\caption{CNN seed variance across the three in-distribution scenarios:
mean $\pm$ population standard deviation over 3 seeds, Ultralytics-native
validation protocol. This protocol differs from the uniform
\texttt{pycocotools} protocol used in Tables~\ref{tab:rgb-lab},
\ref{tab:rgb-ind} and \ref{tab:polar-multimap}, so the seed-0 entries here
are different numbers from the seed-0 rows in those tables for the same
checkpoints; the ranges reported per scenario in the text apply to that
scenario's rows only, not to the table as a whole. These figures
characterize variance within the CNN family and are not comparable across
families; see Tables~\ref{tab:rgb-lab}, \ref{tab:rgb-ind} and
\ref{tab:polar-multimap} for the protocol-matched comparison against the
transformers, whose in-distribution results are single-seed
(Sec.~\ref{sec:training}). Nothing is bolded, since no architecture is
reliably distinguishable from the others at these standard deviations.}
\label{tab:seed-variance}
\vspace{6pt}
\begin{tabular}{lcccc}
\toprule
& \multicolumn{2}{c}{Test} & \multicolumn{2}{c}{Validation} \\
\cmidrule(lr){2-3}\cmidrule(lr){4-5}
Model & mask mAP\textsubscript{50} & mAP\textsubscript{50--95}
      & mask mAP\textsubscript{50} & mAP\textsubscript{50--95} \\
\midrule
\multicolumn{5}{l}{\textit{Controlled RGB}} \\
YOLOv11-n & $0.858 \pm 0.009$ & $0.503 \pm 0.020$ & $0.869 \pm 0.013$ & $0.517 \pm 0.025$ \\
YOLOv11-s & $0.860 \pm 0.018$ & $0.504 \pm 0.013$ & $0.849 \pm 0.007$ & $0.521 \pm 0.007$ \\
YOLOv11-m & $0.865 \pm 0.012$ & $0.510 \pm 0.014$ & $0.844 \pm 0.014$ & $0.514 \pm 0.005$ \\
YOLOv8-s  & $0.779 \pm 0.034$ & $0.463 \pm 0.026$ & $0.855 \pm 0.006$ & $0.549 \pm 0.009$ \\
\addlinespace
\multicolumn{5}{l}{\textit{Industrial RGB}} \\
YOLOv11-n & $0.477 \pm 0.051$ & $0.201 \pm 0.017$ & $0.521 \pm 0.016$ & $0.268 \pm 0.022$ \\
YOLOv11-s & $0.428 \pm 0.039$ & $0.195 \pm 0.019$ & $0.496 \pm 0.032$ & $0.228 \pm 0.010$ \\
YOLOv11-m & $0.221 \pm 0.050$ & $0.081 \pm 0.021$ & $0.372 \pm 0.060$ & $0.193 \pm 0.028$ \\
YOLOv8-s  & $0.440 \pm 0.050$ & $0.226 \pm 0.036$ & $0.493 \pm 0.025$ & $0.242 \pm 0.005$ \\
\addlinespace
\multicolumn{5}{l}{\textit{PolarSens multi-map}} \\
YOLOv11-n & $0.926 \pm 0.021$ & $0.570 \pm 0.012$ & $0.825 \pm 0.028$ & $0.432 \pm 0.009$ \\
YOLOv11-s & $0.890 \pm 0.012$ & $0.545 \pm 0.016$ & $0.838 \pm 0.013$ & $0.442 \pm 0.005$ \\
YOLOv11-m & $0.897 \pm 0.028$ & $0.555 \pm 0.030$ & $0.825 \pm 0.012$ & $0.445 \pm 0.007$ \\
YOLOv8-s  & $0.889 \pm 0.023$ & $0.537 \pm 0.028$ & $0.833 \pm 0.023$ & $0.445 \pm 0.005$ \\
\bottomrule
\end{tabular}
\end{table*}

\subsection{Polarimetric channel fusion}
\label{sec:res-fusion}

Table~\ref{tab:polar-fusion} reports the two fusion variants (nano scale,
trained on the alignment-preserving augmented dataset of
Sec.~\ref{sec:polstrategies}), evaluated with a dedicated multimodal
validator that mirrors the training-time dataset construction. Both
variants perform far below the multi-map baseline of
Table~\ref{tab:polar-multimap} (test mask mAP\textsubscript{50--95}
0.03--0.04 versus a 0.54--0.57 three-seed mean, Table~\ref{tab:seed-variance}),
and below the single-map baseline of
Sec.~\ref{sec:res-polar} as well. Explicit channel fusion is therefore not
competitive with treating the six polarization maps as independent
training samples on this dataset. We attribute this to the combination of
a small training set ($\sim$47 welds) with architectures of
substantially larger effective capacity (a channel-split layer plus, for
the multi-backbone variant, six parallel backbones): both training curves
plateau within the first 5--10 epochs, well before the 30--50-epoch
budget that suffices for the multi-map CNNs, indicating an optimization
regime dominated by insufficient data rather than by insufficient
training time. The primary takeaway remains robust: even with a corrected architecture and a
leakage-free split, both fusion variants perform well below the unmodified
multi-map approach. In contrast, the relative ranking between the two fusion
variants varies across metrics (multi-backbone leads only on test
mAP\textsubscript{50}, while single-backbone wins on the remaining three), so
we draw no firm conclusion on their ordering.

\begin{table*}[t]
\centering
\footnotesize
\setlength{\tabcolsep}{3pt}
\caption{Polarimetric channel fusion (YOLOv11-n backbone, single class),
alignment-preserving augmentation, leakage-free splits (test: 5 unseen welds
/ 30 images; validation: 15 unseen welds / 90 images). Multi-backbone leads
only on test mask mAP\textsubscript{50} and trails on every other column,
including its own validation split; with only 5 test and 15 validation welds,
single high-IoU-threshold hits or misses swing these metrics
disproportionately, so the two variants should be read as comparably poor
rather than one reliably beating the other (see the small-test-set caveat in
Sec.~\ref{sec:discussion}).}
\label{tab:polar-fusion}
\vspace{6pt}
\begin{tabular}{lcccc}
\toprule
& \multicolumn{2}{c}{Test} & \multicolumn{2}{c}{Validation} \\
\cmidrule(lr){2-3}\cmidrule(lr){4-5}
Model & mask mAP\textsubscript{50} & mAP\textsubscript{50--95}
      & mask mAP\textsubscript{50} & mAP\textsubscript{50--95} \\
\midrule
Single-backbone (18ch, split layer) & 0.117 & \textbf{0.040} & \textbf{0.178} & \textbf{0.068} \\
Multi-backbone (6 parallel) & \textbf{0.216} & 0.029 & 0.083 & 0.039 \\
\bottomrule
\end{tabular}
\end{table*}

\subsection{Architectural comparison}
\label{sec:res-arch}

Once evaluated under a unified protocol (Sec.~\ref{sec:protocol}) with quantified seed variance, the cross-model comparison is far less straightforward than single-run metrics imply, for two reasons.
First, within the CNN family, model capacity shows no reliable
in-distribution ranking once seed variance is accounted for
(Table~\ref{tab:seed-variance}: three-seed means for the four architectures fall
within roughly one to two standard deviations of each other in nearly
every scenario and split). The one exception robust to seed noise is the
medium model, which is the worst CNN by a margin exceeding its own
variance on industrial RGB and is never the best elsewhere; beyond that,
we do not consider a nano-versus-small-versus-YOLOv8-s ranking
established.
Second, the transformers do not follow a single pattern relative to the
CNNs. In-distribution, RF-DETR is competitive to outright best in every
scenario under the corrected protocol (best model overall on industrial
RGB, best on controlled-RGB test, tied with the best CNN on PolarSens
mask mAP\textsubscript{50}), while trailing the CNN cluster on the
stricter mAP\textsubscript{50--95} on PolarSens and, on validation splits,
falling back into the CNN range rather than leading it; Mask2Former
remains the weakest model on PolarSens and controlled RGB but is
competitive on industrial RGB test. Latency still tracks capacity cleanly
(Table~\ref{tab:models}): the nano CNN is the fastest model benchmarked
(24\,ms/img) and Mask2Former the slowest (454\,ms/img, roughly
19$\times$ slower), independently of which model is most accurate in a given
scenario. The strongest, most general pattern lies not in in-distribution capacity, but in behavior under domain shift. Across all close-range cross-evaluations, both transformers, particularly RF-DETR, outperform every CNN baseline by a wide margin. The next section analyzes this finding alongside a resolution-matched ablation to separate architectural effects from resolution artifacts.

\section{Discussion and conclusions}
\label{sec:discussion}

\paragraph{Acquisition quality is a system component}
The largest performance factor in this study is not architectural.
Three-seed CNN test means span 0.78--0.87 mask mAP\textsubscript{50} in
controlled conditions and fall to
0.22--0.48 on shop-floor imagery (Table~\ref{tab:seed-variance}), a gap far
wider than any difference between architectures within either scenario. The
transformers degrade less steeply: on the seed-0 checkpoints Mask2Former
drops from 0.610 to 0.523 and RF-DETR from 0.900 to 0.619, both smaller
relative losses than the weakest CNNs, which anticipates the domain-shift
behaviour discussed below. For industrial adoption, the acquisition setup
(background, illumination, and camera distance) deserves at least as much
engineering attention as the choice of network.

\paragraph{Polarimetric imaging for reflective surfaces}
Polarization-derived maps are informative for weld segmentation on
reflective metal. The multi-map strategy with alignment-preserving
augmentation reaches a three-seed mean mask mAP\textsubscript{50} of 0.93
(YOLOv11-n, Table~\ref{tab:seed-variance}) with an unmodified architecture.
Two factors drive this: the complementary information across the six maps
(0.40 to 0.70 mAP\textsubscript{50} from single- to multi-map) and the
geometric augmentation (0.70 to 0.93). Boundary precision remains the open
problem, with mAP\textsubscript{50--95} below 0.58. Explicit channel fusion
does not help: both fusion variants fall well below the multi-map baseline
(Sec.~\ref{sec:res-fusion}), so at this dataset size feeding independent
maps to an unmodified network uses the polarimetric signal more efficiently
than architectural fusion.
Across scenarios, polarimetric multi-map (0.93) is on par with the best
controlled-RGB result (RF-DETR, 0.900, Table~\ref{tab:rgb-lab}), a
difference within the seed variance seen elsewhere in this study (0.01--0.03
on comparable splits). We therefore do not claim polarimetric imaging is
superior to RGB under controlled acquisition. Its practical advantage is
over uncontrolled RGB, where it reaches 0.89--0.93 against 0.25--0.62 for
RGB models (seed-0, Table~\ref{tab:rgb-ind}): it matches controlled-RGB
accuracy without the background, lighting, and distance control that RGB
requires, at the cost of a dedicated camera and polarized illumination.

\paragraph{Model capacity and small data}
Within the CNN family, model capacity gives no reliable in-distribution
ranking once seed variance is accounted for. Across PolarSens and both RGB
scenarios (Table~\ref{tab:seed-variance}), the four architectures' three-seed means fall
within one to two standard deviations of each other in almost every split,
so nano, small, and YOLOv8-s cannot be separated with three seeds and
8--15-image test sets. The one ranking that survives is that the medium
model is the worst CNN on industrial RGB by more than its own noise and is
never best elsewhere. The defensible conclusion is narrow: capacity beyond
the nano/small range brings no measurable gain at this data scale, and the
choice among small variants should rest on latency and memory rather than
an expected accuracy edge. For practice, embedded-friendly models are a
safe default in-distribution, and effort is better spent on data collection
and acquisition control than on model size.

\begin{figure*}[htbp]
\centering
\includegraphics[width=\linewidth]{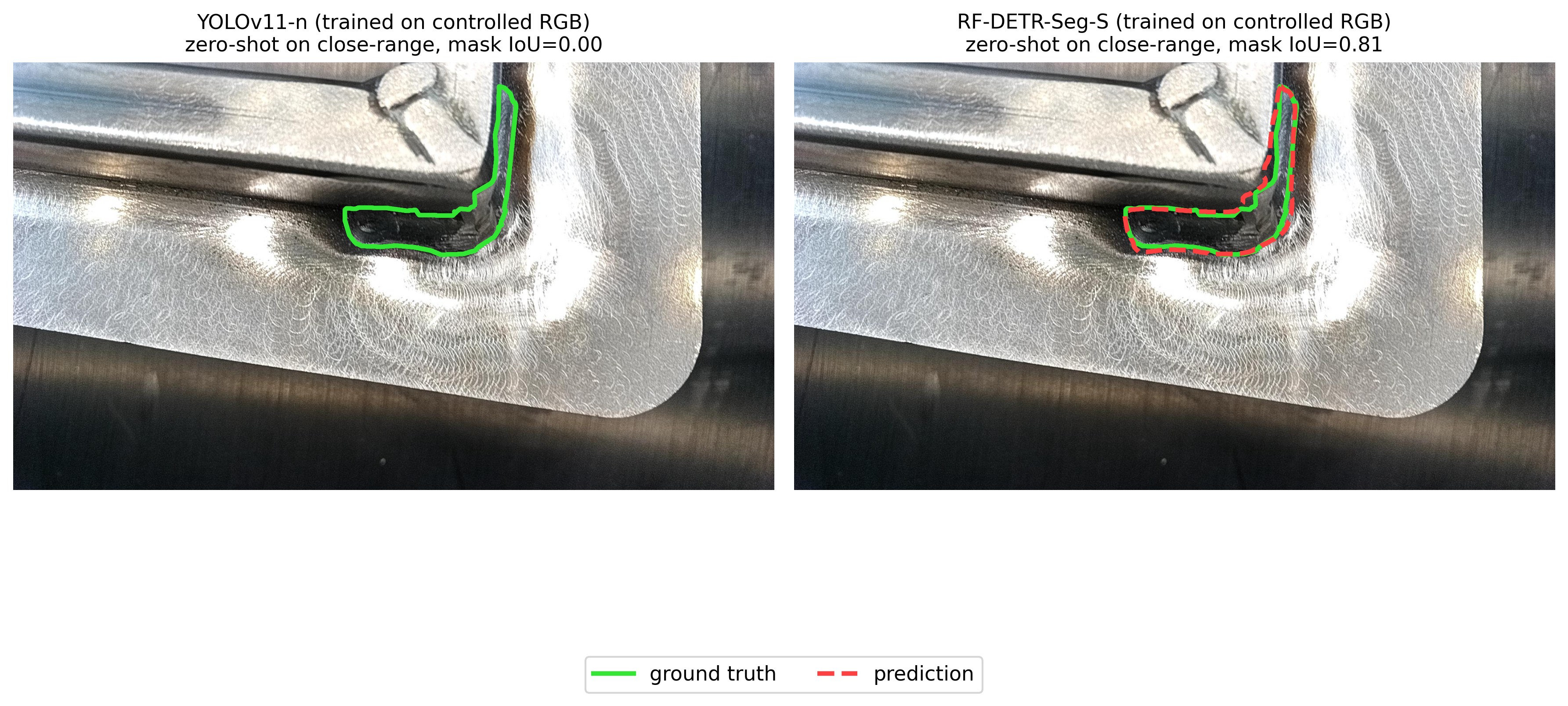}
\caption{Domain-shift comparison on a single close-range test image
(ground truth in solid green, prediction in dashed red), both models
trained on controlled RGB only and evaluated zero-shot: YOLOv11-n produces
no prediction at all, while RF-DETR-Seg-S localizes the weld closely.}
\label{fig:domain-shift}
\end{figure*}

\paragraph{Transformers generalize better under domain shift}
The clearest exception to ``smaller is enough'' is transformer behaviour
under distribution shift. In both zero-shot cross-evaluations to the
close-range set, from controlled RGB (Table~\ref{tab:closerange-lab}) and
from industrial RGB (Table~\ref{tab:closerange-ind}), RF-DETR leads by a wide
margin regardless of training scenario (three-seed mean mask
mAP\textsubscript{50} 0.842 and 0.810, against CNN means of 0.08--0.17 and
0.04--0.19 mask mAP\textsubscript{50}), and its worst seed still beats the
best CNN seed in both cases. The effect does not depend on clean training
data and is not a single-seed artifact. Figure~\ref{fig:domain-shift} shows
it on one image: the controlled-RGB CNN misses the weld entirely, while
RF-DETR trained on the same data localizes it.

Two alternative explanations were tested. The first is
the evaluation threshold: query-based transformers emit low-confidence
detections, and the fixed \texttt{pycocotools} protocol
(Sec.~\ref{sec:protocol}) scores every model over the full precision--recall
curve, removing the bias that a higher cutoff would place on the
transformers. The second is input resolution (Table~\ref{tab:models}: CNNs at
native resolution, transformers at 1120/1024\,px). Because close-range welds
are large in pixels, a lower resolution could in principle favour the
transformers. The ablation in Table~\ref{tab:resolution-ablation} rules this
out: retraining YOLOv11-n from scratch at the transformer resolution
(1120\,px) recovers almost nothing on the close-range set (0.174 mask
mAP\textsubscript{50}), far below RF-DETR at the same resolution (0.842 mask
mAP\textsubscript{50}). The large recovery seen when a natively-trained CNN is
merely run at 1120\,px (0.611 mask mAP\textsubscript{50}) is a side effect of
testing outside the trained resolution, not evidence that resolution explains
the gap.

\begin{table*}[t]
\centering
\footnotesize
\setlength{\tabcolsep}{4pt}
\caption{Resolution ablation: YOLOv11-n trained on controlled RGB,
evaluated zero-shot on the close-range test set (62 images), uniform
\texttt{pycocotools} protocol. RF-DETR-Seg-S is included as the transformer
reference point at the same test resolution. The native-train/test and
RF-DETR rows reproduce values already reported in
Table~\ref{tab:closerange-lab}.}
\label{tab:resolution-ablation}
\vspace{6pt}
\begin{tabular}{lcccc}
\toprule
Configuration & Train (px) & Test (px) & mask mAP\textsubscript{50} & mAP\textsubscript{50--95} \\
\midrule
YOLOv11-n, native (baseline)      & 1920 & 1920 & 0.119 & 0.068 \\
YOLOv11-n, inference-only         & 1920 & 1120 & 0.611 & 0.330 \\
YOLOv11-n, retrained matched      & 1120 & 1120 & 0.174 & 0.081 \\
RF-DETR-Seg-S (reference)         & 1120 & 1120 & \textbf{0.842} & \textbf{0.537} \\
\bottomrule
\end{tabular}
\end{table*}

One feature of the industrial-trained RF-DETR needs care: its close-range
score exceeds its own in-distribution score (0.810 versus a 0.612 three-seed
mean mask mAP\textsubscript{50}, on every seed). A uniformly easier test set
would raise all models, yet the CNNs collapse on the same images, so the
close-range set is not simply easy; RF-DETR transfers to the new viewpoint
well enough to exceed its noisier in-distribution result. Mask2Former does not
show this. Its close-range mean (0.461 mask mAP\textsubscript{50}) is inflated
by a single strong seed (0.693 against roughly 0.34 for the other two, std
0.164), and the pattern is not consistent across runs. We therefore attribute
the domain-shift advantage to RF-DETR specifically, not to transformers as a
class, and read Mask2Former's spread as instability of its head when trained
on only 35 industrial images.
The mechanism most consistent with these results is that the
attention-based, DINOv2- or Swin-pretrained backbones generalize to a large
distance and viewpoint change more gracefully than the CNN backbones.
Confirming this would require an ablation separating backbone pretraining
from architecture family, and a CNN pretrained on a comparably large
corpus, which we did not have. In practice, the result argues for testing
candidate models at the deployment viewpoint rather than defaulting to a
lightweight CNN, especially for a robot-mounted camera whose distance and
angle change along the inspection path.

\paragraph{Limitations}
This is an exploratory study. Test sets are small (5--15 welds per
scenario), so single-weld outcomes can move per-split metrics even after
averaging three seeds; per-weld variance reporting is planned. Three-seed
repetition covers the domain-shift cross-evaluation for both families, but
the in-distribution transformer results (Tables~\ref{tab:rgb-lab},
\ref{tab:rgb-ind}, \ref{tab:polar-multimap}) are single-run, so the smaller
transformer--CNN margins there, such as RF-DETR's in-distribution lead on
industrial RGB, warrant more caution than the domain-shift result.
Mask2Former's close-range variance (std up to 0.164), larger than any other
result here, is itself a finding on head stability in this regime. The RGB
datasets cover only three physical specimens; the resolution ablation
covers one architecture and one scenario; transformer training is capped at
reduced resolution by GPU memory; and the work addresses weld localization
only, not defect classification, which is the natural next step. Finally,
all latency figures are offline measurements on a desktop GPU
(Table~\ref{tab:models}); online validation on live streams and embedded
hardware, including the accuracy--throughput trade-off at the lower input
resolutions such targets require, is left to future work.

\paragraph{Industrial implications}
The results support a phased path. Lightweight CNNs on controlled-acquisition
stations can already assist operators in weld localization, standardizing
inspection and reducing its time. Polarimetric imaging is a worthwhile
upgrade for reflective surfaces, mainly because it removes the
acquisition-control burden rather than because it raises an
already-competitive accuracy ceiling, at the cost of a dedicated camera and
polarized illumination. The offline latency of the fastest model
(24\,ms/img on a desktop GPU) suggests embedded deployment in
robot-guided cells is plausible but unverified on embedded hardware. In all
cases the deployment viewpoint should be checked in advance, given the
domain-shift results.


\section*{Data availability}
Data will be made available on request.

\section*{Acknowledgements}
This work was supported by internal institutional funding.

{
    \small
    \bibliographystyle{ieeenat_fullname}
    \bibliography{cas-refs}
}

\end{document}